\documentclass{article}
\usepackage{spconf,amsmath,graphicx}
\usepackage{hyperref}
\usepackage[inline]{enumitem}

\usepackage{graphicx}
\usepackage{amsthm}
\usepackage{amsmath}
\usepackage{amssymb}
\usepackage{color}
\usepackage{parskip}
\usepackage{mwe}
\usepackage[utf8]{inputenc}

\usepackage{tikz}
\usepackage{pgfplotstable}
\usepackage{makecell}
\pgfplotsset{compat=1.14}
\usepgfplotslibrary{statistics}
\usepgfplotslibrary{groupplots}

\title{Deep geometric knowledge distillation with graphs}
\newlist{inlinelist}{enumerate*}{1}
\setlist*[inlinelist,1]{%
  label=(\roman*),
}

\name{\normalsize{Carlos Lassance$^{\diamond}$, Myriam Bontonou$^{\diamond \dagger}$, Ghouthi Boukli Hacene$^{\diamond \dagger}$, Vincent Gripon$^{\diamond}$, Jian Tang$^{\circ \dagger}$, Antonio Ortega$^{\wedge}$}
\thanks{This work was supported in part by the Brittany region. Computations were performed using a Titan-V, courtesy of NVIDIA.}}

\address{$^{\diamond}$ IMT Atlantique, Lab-STICC, $^{\wedge}$ University of Southern California \\ $^{\dagger}$ Mila, $^{\circ}$ HEC Montréal \\}
\begin{document}
\ninept
\maketitle
\begin{abstract}
In most cases deep learning architectures are trained disregarding the amount of operations and energy consumption. However, some applications, like embedded systems, can be resource-constrained during inference. A popular approach to reduce the size of a deep learning architecture consists in distilling knowledge from a bigger network (teacher)  to a smaller one (student). Directly training the student to mimic the teacher representation can be effective, but it requires that both share the same latent space dimensions. In this work, we focus instead on relative knowledge distillation (RKD), which considers the geometry of the respective latent spaces, allowing for dimension-agnostic transfer of knowledge. Specifically we introduce a graph-based RKD method, in which graphs are used to capture the geometry of latent spaces. Using classical computer vision benchmarks, we demonstrate the ability of the proposed method to efficiently distillate knowledge from the teacher to the student, leading to better accuracy for the same budget as compared to existing RKD alternatives.
\end{abstract}

\section{Introduction}

Deep Neural Networks (DNNs) have been shown to outperform other machine learning methods in numerous tasks~\cite{zagoruyko2016wide,wu2016google}. Their success is heavily linked to the availability of large amounts of data and special purpose hardware, e.g., graphics processing units (GPUs) allowing significant levels of parallelism. However, this need for a significant amount of computation is a limitation in the context of embedded systems, 
where energy and memory are constrained. As a result, numerous recent works~\cite{courbariaux2015binaryconnect,han2015deep} have focused on compressing deep learning architectures so that the inference process can be run on embedded devices. 

An approach to reduce the size of a deep learning architecture is individual knowledge distillation (IKD)~\cite{ba2014deep,HintonKD,fitnets}, where the basic idea is to use an available large network, called {\em teacher}, to train a smaller one, called {\em student}, in an attempt to reduce the loss of accuracy in replacing the former by the latter. Initial IKD techniques~\cite{HintonKD} focused on using the output representations of the teacher as a target for the smaller architecture, i.e., the student is trained to mimic the teacher decisions. As a consequence, the knowledge acquired while training the teacher is diffused throughout all layers of the student during backpropagation. More recent works have reached better accuracy by performing this process layer-wise, or block-wise for complex architectures~\cite{fitnets,LIT}.
However, IKD can be directly performed layer-wise only if the student and the teacher have inner data representations with the same dimension~\cite{LIT}, which narrows down significantly the  pairs (teacher, student) to which this can be applied. In an attempt to overcome this limitation, in~\cite{fitnets} the authors introduced extra layers meant to perform distillation during training. These layers are then disregarded during inference. This is problematic as the extra layers encode part of the knowledge distilled by the teacher.

Forcing architectures for teacher and student to have the same latent space dimensions is not practical. Indeed, in~\cite{EfficientNet} the authors show that for efficiently scaling down neural networks one should consider three main aspects: \begin{inlinelist} \item network depth (number of layers); \item network width (number of feature maps per layer); \item resolution (size of the input).\end{inlinelist} Note that the two latest points are related to dimension of latent spaces. In an effort to allow distillation to be performed layer-wise on architectures with varying dimensions, recent works~\cite{TripledDistill, RKD} have introduced distillation in a dimension-agnostic manner. To do so, these methods focus on the relative distances of the intermediate representations of training examples, rather than on the exact positions of each example in their corresponding domains. These methods are referred to as relational knowledge distillation (RKD).

In the present work we extend this notion of RKD by introducing {\em graph knowledge distillation} (GKD). As in our prior work~\cite{griort}, we construct graphs where vertices represent training examples, and the edge weight between two vertices is a function of the similarity between the representations of the corresponding  examples at a given layer of the network architecture. The main motivation for this choice is that even though representations generally have different dimensions in each architecture, the size of the corresponding graphs is always the same  (since the number of nodes is equal to the number of training examples). Thus, information from graphs generated from the teacher architecture can be used to train the student architecture by introducing a discrepancy loss between their respective adjacency matrices during training.
Our main contributions are: we introduce a layer-wise distillation process using graphs, extending the RKD framework, and we demonstrate that this method can improve the accuracy of students trained in the context of distillation, using standard vision benchmarks. The reported gains are about twice as important as those obtained by using standard RKD instead of no distillation.

The paper is organized as follows: In Section~\ref{related_work}, we present related work. In Section~\ref{methodology} we define notations, introduce our proposed framework and detail the generalizations it allows. In Section~\ref{experiments}, we perform experiments and discuss them. Section~\ref{conclusion} provides some conclusions.

\section{Related work}
\label{related_work}

\textbf{Neural network compression:} Reducing DNNs size and computational power is an active field of research that attracted a lot of attention since it eases implementation of DNNs on resource-limited devices such as smartphones, enabling mobile applications. Some authors propose to use high level approaches such as pruning techniques~\cite{yu2018nisp,huang2018learning}, factorization~\cite{han2015deep,wu2018deep}, efficient neural network architectures and/or layers~\cite{EfficientNet,sandler2018mobilenetv2,ma2018shufflenet,hacene2019attention}, knowledge distillation~\cite{ba2014deep,HintonKD,fitnets,LIT,TripledDistill,TripletLoss,stock2019and} and quantizing weights and activations~\cite{courbariaux2015binaryconnect,wu2018deep,stock2019and}. All these approaches can be seen as complementary with each other (e.g. ~\cite{stock2019and} combines distillation and weight quantization.). Our work is better defined as neural network compressing via distillation and is therefore complementary to the mentioned approaches.

\textbf{Neural network distillation:} Following~\cite{RKD}, we distinguish approaches transferring knowledge input by input, from approaches focusing on relative distances on a batch of inputs. The former are known as individual knowledge distillation (IKD)~\cite{ba2014deep,HintonKD,LIT,fitnets,stock2019and} and the latter as relational knowledge distillation (RKD)~\cite{TripledDistill, RKD}.

In IKD, each example is treated independently, which means that the transferred representations from teacher to student have to be of the same dimension. This is not a problem if we consider only the output of the network~\cite{HintonKD}, but it has been shown that by doing it in layer-wise/block-wise fashion it is possible to get better results. To deal with this, \cite{fitnets} adds linear mappings to the student network so that its representations match those of the teacher in terms of dimensions, while~\cite{LIT} proposes to do it only at the end of each block, reducing the amount of parameters inside the block, but keeping the same dimensions for the output.

On the other hand, RKD considers relative positioning of examples in latent spaces, and then compare these between teacher and student. It is therefore dimension agnostic. In~\cite{TripledDistill} the authors are inspired by the triplet loss~\cite{TripletLoss}. In another vein, in~\cite{RKD} a general framework is introduced, using either the Euclidean distance between pairs of examples, or angular distance between triplets. Our work can be seen as an extension of the Euclidean version of RKD, explicitly using graphs to model the relational distances. This allows us to derive more diverse variations of the method, such as higher-order geometric relations and graph signal analysis, while also being able to retrieve the baselines introduced in~\cite{RKD}.

\textbf{Graphs and neural networks:} The use of graphs in neural networks has been of high interest to the community, thanks to the developments in Graph Signal Processing (GSP)~\cite{shuman2013emerging}. Most works are interested in dealing with inputs defined on graphs~\cite{kipf2016semi}. Other works use graphs as a proxy to the topology of intermediate representations of inputs within the network. They are then used to interpret what the network is learning~\cite{griort,anirudh2017margin} or to enhance its robustness~\cite{lassance2018laplacian,svoboda2018peernets}. Based on our prior work~\cite{griort}, in this work we extend the concept of using graphs to represent the geometry of latent spaces in order to perform relational knowledge distillation.

\section{Methodology}
\label{methodology}

\subsection{IKD and RKD}

Let  $T$ and  $S$ denote the teacher and student architectures, respectively. We aim at transferring knowledge from $T$ to $S$, where $S$ typically contains fewer parameters than $T$. For presentation simplicity, we assume that both architectures generate the same number of inner representations, even though the method could easily be extended to cases where this is not true. In the context of distillation, we consider that the teacher has already been trained, and that we want to use both the training set and the inner representations of the teacher in order to train the student. This is an alternative to directly training the student using only the training data (which we refer to as ``baseline'' in our experiments). Formally, we use the following loss to train the student:
\begin{equation}\mathcal{L} = \mathcal{L}_\text{task} + \lambda_{\text{KD}} \cdot \mathcal{L}_\text{KD}\;,\label{loss}\end{equation}
where $\mathcal{L}_\text{task}$ is typically the same loss that was used to train the teacher (e.g. cross-entropy), $\mathcal{L}_\text{KD}$ is the distillation loss and $\lambda_{\text{KD}}$ is a scaling parameter to control the importance of the distillation with respect to that of the task. 

Denote $X$ a batch of input examples and  $\Lambda$ a set of layers on which we aim at transferring knowledge. When processing an input $\mathbf{x}$, a deep neural network architecture $A$ generates a series of inner representations, one for each layer $\ell$ of the network: $\mathbf{x}^A_1, \mathbf{x}^A_2,\dots,\mathbf{x}^A_\ell,\dots,\mathbf{x}^A_L$.
IKD approaches try to directly compare the inner representations of both teacher and student when processing the same input $\mathbf{x}$. Thus, the IKD loss can be written as~\cite{ba2014deep,HintonKD,fitnets,LIT,stock2019and}:
$$\mathcal{L}_\text{IKD} = \sum_{\ell\in \Lambda}\sum_{\mathbf{x}\in X}{\mathcal{L}_d(\mathbf{x}^S_\ell,\mathbf{x}^T_\ell)},$$
where, typically, $\mathcal{L}_d$ is a measure of the distance between its arguments, which requires that they have the same dimension.

In contrast, RKD approaches consider relative metrics between the respective inner representations of the networks to be compared. In the specific case of RKD-D~\cite{RKD}, the mathematical formulation becomes: 
$$\mathcal{L}_\text{RKD-D} = \sum_{\ell\in \Lambda}\sum_{(\mathbf{x},\mathbf{x'})\in X^2}{\mathcal{L}_d\left(\frac{\|\mathbf{x}^S_\ell-\mathbf{x'}^S_\ell\|_2}{\Delta^S_\ell},\frac{\|\mathbf{x}^T_\ell - \mathbf{x'}^T_\ell\|_2}{{\Delta^T_\ell}}\right)},$$
where $\Delta^A_\ell$ is the average distance between all couples $(\mathbf{x}^A_\ell,\mathbf{x'}^A_\ell)$ for the given architecture at layer $\ell$ and $\mathcal{L}_d$ is the Huber loss~\cite{huber1992robust}.
The main advantage of using RKD is that it allows to distillate knowledge from an inner representation of the teacher to one of the student, even if their respective dimensions are different.

\subsection{Proposed Approach: Graph Knowledge Distillation (GKD)}

Instead of directly trying to make the distances between data points in the student match those of the teacher, we consider the problem from a graph perspective.
Given an architecture $A$, a batch of inputs $X$ and a layer $\ell$, we compute the corresponding inner representations $X^A_\ell = [\mathbf{x}^A_\ell, \mathbf{x} \in X]$. Using a given similarity metric, we can then use these representations to define a $k$-nearest neighbor similarity graph $\mathcal{G}^A_\ell(X) = \langle X^A_\ell, \mathbf{W}^A_\ell\rangle$. The graph contains a node for each input in the batch, and the edge weight $\mathbf{W}^A_\ell[ij]$ represents the similarity between the $i$-th  and the $j$-th elements of $X^A_\ell$, or 0 (depending on $k$). In this work, we use the cosine similarity. To avoid giving excessive importance to outliers, we also normalize the adjacency matrix as follows: $\mathbf{A}^A_\ell \triangleq \mathbf{D}^{-1/2} \mathbf{W}^A_\ell \mathbf{D}^{-1/2}$, 
where $\mathbf{D}$ is the diagonal degree matrix of the graph.

While training the student, we input our training batch into both the student architecture and the (now fixed) previously trained teacher architecture. This provides a similarity graph for each layer in $\Lambda$. The loss we aim to minimize combines the task loss, as expressed in Equation~(\ref{loss}), with the following graph knowledge distillation (GKD) loss: 
\begin{equation}\mathcal{L}_{\text{GKD}} = \sum_{\ell\in \Lambda}{\mathcal{L}_d(\mathcal{G}^S_\ell(X),\mathcal{G}^T_\ell(X))}\;.\label{belowa}\end{equation}

In our work, we mainly consider the case where $\mathcal{L}_d$ is the $\mathcal{L}_2$ distance between the adjacency matrix of its arguments.

The GKD loss measures the discrepancy between the adjacency matrices of teacher and student graphs. In this way the geometry of the latent representations of the student will be forced to converge to that of the teacher. Our intuition is 
that since the teacher network is expected to generalize well to the test, mimicking its latent representation geometry should allow for better generalization of the student network as well.


As the values of $\mathbf{\mathcal{A}}$ are normalized by the degree matrix, the Huber loss simplifies to the square of the Frobenius norm of the matrix resulting from the difference of the student and teacher adjacency matrices. An equivalent definition of our proposed loss is:
\begin{equation}\mathcal{L}_{\text{GKD}} = \sum_{\ell \in \Lambda}{\|\mathbf{A}^S_\ell-\mathbf{A}^T_\ell\|_2^2}\;.\label{below}\end{equation}

A first obvious advantage of GKD with respect to RKD-D is the fact it allows normalization over the batch of inputs, yielding to a more robust process. This is discussed in Section~\ref{normalization}.
Amongst other degrees of freedom that are available to us, in this paper we focus on three possible variations of the method:
\begin{enumerate}
    \item Locality: varying the value $k$ when constructing $k$-nearest neighbor graphs. This allows us to focus only on the closest neighbors of each example,
    \item Higher order: taking powers $p$ of the normalized adjacency matrix of considered graphs before computing the loss. By considering higher powers of matrices $\mathbf{A}$, we consider higher-order geometric relations between inner representations of inputs,
    \item Task specific: considering only examples of the same (resp. distinct) classes when creating the edges of the graph, thus focusing on the clustering (resp. margin) of classes.
\end{enumerate}

\section{Experiments}
\label{experiments}
We perform two types of experiments. We first evaluate accuracy of RKD-D and proposed GKD using the CIFAR-10 and CIFAR-100 datasets of tiny images~\cite{krizhevsky2009learning}. We then look at proposed variations of GKD. 

\subsection{Hyperparameters}

We train our CIFAR-10/100 networks for 200 epochs, using standard Stochastic Gradient Descent (SGD) with batches of size 128 
($|X|=128$) and an initial learning rate of $0.1$ that is decayed by a factor of $0.2$ at epochs $60$, $120$ and $160$. We also add a momentum of $0.9$ and follow standard data augmentation procedure~\cite{zagoruyko2016wide}. We use a WideResNet28-1~\cite{zagoruyko2016wide} architecture for our teacher network, while the student network uses a WideResNet28-0.5. In terms of scale, WideResNet28-0.5 has approximately 27\% of the operations and parameters of WideResNet28-1. All these architectures are particularly small compared to state-of-the-art. We use a network of same size of the students but trained without a teacher as a baseline that we call Vanilla. Our RKD-D~\cite{RKD} students are trained with the parameters from~\cite{RKD}, $\lambda_\text{RKD-D}=25$ and applied to the output of each block. We applied the same values for GKD. Note that all these choices were made to remain as consistent as possible with existing literature. For each student network we run 3 tests and report the median value. The code for reproducing the experiments is available at~\url{https://github.com/cadurosar/graph_kd}.

\subsection{Direct comparison between GKD and RKD-D}

In a first experiment we simply evaluate the test set error rate when performing distillation. Results are summarized in Table~\ref{errorrate}. We compare student sized networks trained without distillation (Baseline), with GKD and RKD-D~\cite{RKD}. We also report the performance of the teacher (that can be seen as an upper bound). RKD-D~\cite{RKD} by itself provides a small gain in error rate with respect to the Baseline approach, while GKD outperforms RKD-D by almost the same gain.

\begin{table}[ht]
\centering
\caption{Error rate comparison of GKD and RKD-D on the CIFAR-10/100 datasets.}
\vspace{.5cm}
\begin{tabular}{|c|c|c|c|}
\hline
Method                   & CIFAR-10 & CIFAR-100 & Relative size   \\ \hline 
Teacher                  & 7.27\%     & 31.26\%  & 100\% \\ \hline
Baseline                 & 10.26\%   & 38.50\% & 27\%  \\ \hline
\hline
RKD-D~\cite{RKD}         & 10.06\%    & 38.26\% & 27\%  \\ \hline
GKD                      & \textbf{9.70\%}     & \textbf{38.17\%} & 27\%  \\ \hline

\end{tabular}
\label{errorrate}
\end{table}

\subsection{Effect of the normalization}
\label{normalization}

To better understand why GKD performed better than RKD-D we analyze the contribution of each example in a batch in both the GKD loss and the RKD-D one. If our premise from Section 3.2 is correct, by using a degree normalized adjacency matrix instead of the distance pairs directly, most examples will be able to contribute to the optimization. To do so, we compute the respective loss, for each block, using 50 batches of 1000 training set examples and analyze the median amount of examples that are responsible for 90\% of the loss at each block.  In Table~\ref{normalization_table}, we present the results. As we suspected for GKD shows a significant advantage on the number of examples responsible for 90\% of the loss.

\begin{table}[ht]
\centering
\caption{Comparison of the effect of the normalization on the amount of examples that it takes to achieve 90\% of the total loss value.}
\vspace{.5cm}
\begin{tabular}{|c|c|c|}
\hline
Block position in the architecture & RKD-D            & GKD              \\ \hline\hline
Middle & 83.70\%          & \textbf{86.50\%} \\ \hline
Final & 82.05\%          & \textbf{83.60\%} \\ \hline
\end{tabular}
\label{normalization_table}
\end{table}

\subsection{Classification consistency}

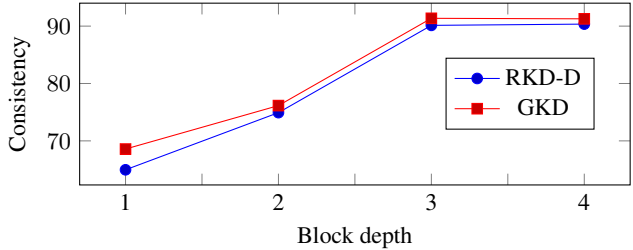
\begin{figure}[ht]
 \begin{center}
   \begin{tikzpicture}
       \begin{axis}[
           xlabel=Block depth,
           ylabel=Consistency,
           legend style={at={(0.8,0.7)},anchor=north},
           width=.5\textwidth,
           height=4cm,
           xticklabels={, ,1, ,2, ,3, ,4}
]
         \addplot table {consistency/RKD-D.txt};
         \addlegendentry{RKD-D}
         \addplot table {consistency/GKD.txt};
         \addlegendentry{GKD}
       \end{axis}
   \end{tikzpicture}
   \vspace{-.3cm}
 \end{center}
    \caption{Analysis of the consistency of classification compared to the teacher, across layers of RKD-D and GKD students. We consider the output of the network as the ``fourth block''.}
    \label{consistency}
    \vspace{-15pt}
\end{figure}

We now take our trained students and compare their outputs to the trained teacher's outputs. For the output of each WideResNet block we compute the classification of a simple Logistic Regression, while the network's final output is already a classifier. The ideal scenario would be one where the student is 100\% consistent with the teacher's decision on the test set, as this would greatly improve the classification performance when compared to the baseline. The results are depicted in Figure~\ref{consistency}. As expected the GKD was able to be more consistent with the teacher than the RKD-D.

\subsection{Spectral analysis}

\begin{figure}[ht]
 \begin{center}
   \begin{tikzpicture}
       \begin{axis}[
           xlabel=Block depth,
           ylabel=Smoothness,
           title=Teacher's Fiedler vector,
           width=.24\textwidth,
           height=4cm]
         \addplot table {fiedler/RKD-D.txt};
         \addlegendentry{RKD-D}
         \addplot table {fiedler/GKD.txt};
         \addlegendentry{GKD}
       \end{axis}
   \end{tikzpicture}
   \begin{tikzpicture}
       \begin{axis}[
           xlabel=Block depth,
           title=Label binary indicator signal,
           width=.24\textwidth,
           height=3.95cm]
         \addplot table {signal/RKD-D.txt};
         \addlegendentry{RKD-D}
         \addplot table {signal/GKD.txt};
         \addlegendentry{GKD}
       \end{axis}
   \end{tikzpicture}
 \end{center}
    \caption{Analysis of the smoothness evolution across layers of the RKD-D and GKD students. In the right we have the label binary indicator signal and in the left we use the Teacher's Fiedler vector as a signal.}
    \label{spectral}
    \vspace{-15pt}
\end{figure}
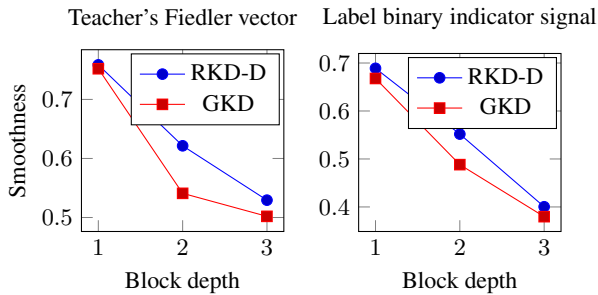

Given that we have introduced intermediate representation graphs, it is quite natural to analyze performance from a graph signal processing perspective~\cite{shuman2013emerging}. We propose to do this by considering specific graph signals $\textbf{s}$ and computing their respective smoothness on each of the two graphs. Smoothness is computed as $\sigma=\textbf{s}^\top\textbf{L}\textbf{s}$, where $L$ is the Laplacian of the studied graph ($L=\textbf{D}-\textbf{W}$). Lower values of $\sigma$ mean that the signal is better aligned with the graph structure. We create graphs with 1000 examples chosen at random from the training set. The signals that we consider are \begin{inlinelist} \item the label binary indicator signal, which we have previously shown to be a good indicator for overfitting/underfitting~\cite{griort} or robustness~\cite{lassance2018laplacian}; \item the Fiedler eigenvectors from each intermediate representation in the teacher, which allow us to compare the clustering of both networks and how they evolve over successive layers.\end{inlinelist} The results are depicted in Figure~\ref{spectral}. We can see that both signals have more smoothness in the graphs generated by GKD. This means that the geometry of the latent spaces from GKD are more aligned to those of the teacher. 

\subsection{Effect of locality}

We now consider variations of the proposed GKD method. The first one is the effect of changing the value $k$. The main effect of lowering the value of $k$ is to focus on most similar examples. Indeed, large distances, typically at early layers, can be meaningless. Results are summarized in Table~\ref{neighbors}. Lower values of $k$ yield better results, showing that it is preferable to focus on the closest neighbors than all distances. This is similar to results such as~\cite{TripletLoss}, where the authors show that it is best to concentrate on the hardest cases, instead of trying to solve all of them.

\begin{table}[ht]
\centering
\caption{Analysis of the effect of varying $k$ on the networks error rates.}
\vspace{.5cm}
\begin{tabular}{|c||c|c|c|}
\hline
$k$        & $|X|$ & $|X|/2$ & 5 \\ \hline
Error Rate & 9.70\% & 9.55\% & \textbf{9.43\%}\\
\hline
\end{tabular}
\label{neighbors}
\end{table}

\subsection{Higher orders}

We then study the effect of varying the power of adjacency matrices $p$. This allows us to consider higher-order geometric relations between inner representation of inputs when compared to fixing $p$ to 1. The results are presented in Table~\ref{powers}. Higher-order geometric relations do not seem to help the transfer of knowledge. One possible reason for this result is that using all the distances for a higher-order relation introduces too much noise.

\begin{table}[ht]
\centering
\caption{Analysis of the effect of varying $p$ on the networks error rates.}
\vspace{.5cm}

\begin{tabular}{|c||c|c|c|}
\hline
$p$        & 1 & 2 & 3 \\ \hline
Error Rate & \textbf{9.70\%} & 10.44\% & 10.37\%\\\hline
\end{tabular}
\label{powers}
\end{table}

\subsection{Task specific graph signals}

Now we evaluate the effects of considering only intra or inter-class distances. If we consider only inter-class distances we can focus mostly on having a similar margin in both teacher and student. On the other hand, considering only intra-class distances would force both networks to perform the same type of clustering on the classes. The results are presented in Table~\ref{signal}. In this case, focusing on the margin helped, while concentrating on the clustering was not effective. This result is similar to what we found in our prior work~\cite{lassance2018laplacian}, which shows that the margin is a better tool to interpret the network results than the class clustering.

\begin{table}[ht]
\centering
\caption{Analysis of the effect of focusing either on the margin or on the class clustering.}
\vspace{.5cm}

\begin{tabular}{|c|c|c|}
\hline
Pairs           & Error Rate     \\ \hline\hline
All possible pairs  & 9.70\%      \\ \hline
Only pairs of distinct classes & \textbf{9.54\%}  \\ \hline
Only pairs of the same class & 10.35\%   \\ \hline
\end{tabular}
\label{signal}
\end{table}

\section{Conclusion}
\label{conclusion}
We introduced graphs knowledge distillation, a method using graphs to transfer knowledge from a teacher architecture to a student one. By using graphs, the method opens the way to numerous variations that can significantly benefit the accuracy of the student, as demonstrated by our experiments. In future work we consider: \begin{inlinelist} \item using more appropriate graph distances, such as in~\cite{bunke1998graph,segarra2015diffusion}; \item doing a more in-depth exploration of how to properly scale the student network, e.g. following~\cite{EfficientNet}; \item combining with approaches such as~\cite{TripletLoss,bontonou2019} to train a teacher network in a layer-wise fashion. \end{inlinelist}

\bibliographystyle{IEEEbib}
\bibliography{refs}
\end{document}